%% file: root.tex
\documentclass[letterpaper, 10 pt, conference]{ieeeconf}  % Comment this line out if you need a4paper

\IEEEoverridecommandlockouts                              % This command is only needed if 
\usepackage{graphicx}
\usepackage{booktabs}
\usepackage{amsmath}  
\usepackage{amsfonts} 
\usepackage{subcaption}
\usepackage{placeins}
\usepackage{makecell}
\usepackage{diagbox}
\usepackage{float}
\usepackage{afterpage}
\usepackage{xcolor}
\usepackage{hyperref}
\usepackage{xspace}
\title{\LARGE \bf
Guardians as You Fall: Active Mode Transition for Safe Falling
}

\author{Yikai Wang, Mengdi Xu, Guanya Shi, and Ding Zhao% <-this % stops a space
\thanks{Authors are associated with Carnegie Mellon University, USA.
{\texttt{wangyiji20@mails.tsinghua.edu.cn}}
{\{\texttt{mengdixu,guanyas,dingzhao\}@andrew.cmu.edu}}}\\
\thanks{Project website and videos: \href{https://sites.google.com/view/guardians-as-you-fall/}{https://sites.google.com/view/guardians-as-you-fall/}}
}

\begin{document}

\newcommand{\guanya}[1]{{\color{blue}[Guanya: #1]}}
\definecolor{mengdi_color}{RGB}{86,151,137}
\hypersetup{
    citecolor = {mengdi_color}
}
\newcommand{\mengdi}[1]{{\color{mengdi_color}[Mengdi: #1]}}

\newcommand{\method}{GYF\xspace} 

\maketitle
\thispagestyle{empty}
\pagestyle{empty}

%%%%%%%%%%%%%%%%%%%%%%%%%%%%%%%%%%%%%%%%%%%%%%%%%%%%%%%%%%%%%%%%%%%%%%%%%%%%%%%%
\begin{abstract}
Recent advancements in optimal control and reinforcement learning have enabled quadrupedal robots to perform various agile locomotion tasks over diverse terrains.
During these agile motions, ensuring the stability and resiliency of the robot is a primary concern to prevent catastrophic falls and mitigate potential damages.
Previous methods primarily focus on recovery policies \emph{after} the robot falls. There is no active safe falling solution to the best of our knowledge. 
In this paper, we proposed Guardians as You Fall (\method), a safe falling/tumbling and recovery framework that can actively tumble and recover to stable modes to reduce damage in highly dynamic scenarios. The key idea of \method~is to adaptively traverse different stable modes via active tumbling \emph{before} the robot shifts to irrecoverable poses. Via comprehensive simulation and real-world experiments, we show that \method~significantly reduces the maximum acceleration and jerk of the robot base compared to the baselines. In particular, \method reduces the maximum acceleration and jerk by $20\%$ $\sim$ $73\%$ in different scenarios in simulation and real-world experiments. \method~offers a new perspective on safe falling and recovery in locomotion tasks, potentially enabling much more aggressive explorations of existing agile locomotion skills.

% Some approaches introduce fall and recovery controllers to minimize damage in the event of a fall. However, previous methods for damage reduction struggle to handle highly dynamic situations, considering the limitations imposed by the robot's structure and actuation forces. 

%\guanya{compared to XXX}.
%\guanya{add some numbers}.
% In this study, we introduce a novel approach to reduce damage in high dynamic situations by through adaptive transitions to and between stable states. We demonstrate that, using the presented method, robots can adaptively shift between stable states, significantly reducing both the contact force on the body and vertical momentum changes. Our approach offers a fresh perspective on damage reduction and has the potential to expand the boundaries of exploration in quadrupedal agile control. 
\end{abstract}

\input{intro}

\input{related}

\input{method}

\input{result}

\input{conclusion}

%%%%%%%%%%%%%%%%%%%%%%%%%%%%%%%%%%%%%%%%%%%%%%%%%%%%%%%%%%%%%%%%%%%%%%%%%%%%%%%%

% \addtolength{\textheight}{-12cm}   % This command serves to balance the column lengths
                                  % on the last page of the document manually. It shortens
                                  % the textheight of the last page by a suitable amount.
                                  % This command does not take effect until the next page
                                  % so it should come on the page before the last. Make
                                  % sure that you do not shorten the textheight too much.

%%%%%%%%%%%%%%%%%%%%%%%%%%%%%%%%%%%%%%%%%%%%%%%%%%%%%%%%%%%%%%%%%%%%%%%%%%%%%%%%

%%%%%%%%%%%%%%%%%%%%%%%%%%%%%%%%%%%%%%%%%%%%%%%%%%%%%%%%%%%%%%%%%%%%%%%%%%%%%%%%

%%%%%%%%%%%%%%%%%%%%%%%%%%%%%%%%%%%%%%%%%%%%%%%%%%%%%%%%%%%%%%%%%%%%%%%%%%%%%%%%

% \clearpage
% \section*{APPENDIX}

% Appendixes should appear before the acknowledgment.
%\newpage
\section*{ACKNOWLEDGMENT}
We would like to thank Shiqi Liu and Xilun Zhang from Safe AI lab for helping with hardware experiments and providing suggestions about the experiments.

%%%%%%%%%%%%%%%%%%%%%%%%%%%%%%%%%%%%%%%%%%%%%%%%%%%%%%%%%%%%%%%%%%%%%%%%%%%%%%%%

\bibliographystyle{unsrt}
\bibliography{root}

\end{document}

%% file: intro.tex
\section{INTRODUCTION}

 % - dynamic robot tasks
 % - safety is important

 % - how other papers deal with falling for Legrobot plus their drawbacks
 %   - bipedal robot
 %   - quadruped 

 % - how our proposed method tackles such drawbacks

 % - the main contribution of our method
 %   - stable modes
 %   - framework
 %   - experiment results better than others

Recent advancements in reinforcement learning (RL) and optimal control have empowered quadrupedal robots to perform a series of dynamic tasks, such as navigating diverse terrains in the wild~\cite{choi2023learning,lee2020learning,kumar2021rma}, achieving high-speed running~\cite{margolisyang2022rapid, ji2022concurrent}, engaging in parkour~\cite{zhuang2023robot}, executing jumps~\cite{yang2023cajun}, and standing up on hind legs~\cite{smith2023learning, cheng2023legs}. 
These tasks involve substantial kinetic or potential energy, heightening the risk of severe falls. 
While current research focuses on enhancing robot safety during agile movements by optimizing the balance capability, risks of falling persist due to sim-to-real gaps, as well as unforeseen disturbances.

% Existing works enhance robots' safety when conducting agile motions by improving their balance capabilities to avoid falling or tumbling. 
% However, the possibility of falling still exists due to sim-to-real gaps and unexpected perturbation. 
% \guanya{I didn't understand the previous sentence. Why don't just say something like ``existing works enhance safety by maximizing the balance capabilities to avoid any falling or tumbling''?} 
% \guanya{due to sim-to-real gaps and XXX}

% \guanya{No transitions between the first two paragraphs. Consider something like: Few works explicitly design falling policies to minimize damage. For bipedal robots,} 

Despite the great significance of ensuring safe falls in legged robots, only a handful of studies specifically aim to design policies that minimize damage during dynamic movements, and most focuses on bipedal robots. 
The safe falling controllers for bipedal robots typically define the motion based on body part contact sequences and guide the robot to a stable mode where other body parts besides the feet are landed on the ground to ensure stability \cite{samy2015falls, samy2017qp, yun2014tripod, ha2015multiple}.
While these controllers actively manage falling behaviors, they rely on a predetermined signal to trigger the controller with action ranges limited by the contact sequence definitions.
Additionally, the bipedal robots are usually initialized in a posture close to a standing mode, and it is unclear about the falling controllers' performance in highly dynamic scenarios.  
Furthermore, recovering to bipedal standing mode after a fall is under-explored in these literatures. 

% They heavily rely on simplified dynamic models 
% and can only generate limited falling behaviors restricted by contact sequence definitions. 
% They typically rely on a pre-defined signal to detect the falling of bipedals from a standing posture, and it is unclear about the controllers' performance in highly dynamic scenarios. 
% The bipedal robots are usually initialized in a posture close to a standing mode, and it is unclear about the falling controllers' performance in highly dynamic scenarios. 

% The safe falling controllers for bipedal robots either follow a predefined contact mode~\cite{samy2015falls, samy2017qp, yun2014tripod} or are optimized on the fly~\cite{ha2015multiple}.
% Such controllers heavily rely on simplified dynamic models and can only generate limited falling behaviors restricted by the contact sequence definitions. They typically rely on a pre-defined signal to detect the falling of bipedals from a standing posture, and it is unclear about the controllers' performance in highly dynamic scenarios. 

In contrast, in quadrupedal robots, most safe falling literature focuses on developing recovery policies after a fall \cite{ma2023learning, lee2019robust} without considering the damage reduction during the falling process. 
Most recent works attach robot arms to quadrupeds to help avoid falling too severely and restore balance~\cite{ma2023learning, tang2023towards}. However, these methods modify the robot's hardware structure, potentially altering the task scope and reducing locomotion agility. To the best of our knowledge, there is no active safe falling solution for quadrupedal robots in dynamic scenarios.
% involve tracking a contact sequence, either following a predefined mode~\cite{samy2015falls,samy2017qp, yun2014tripod} or optimizing online~\cite{ha2015multiple} or offline~\cite{8460500}. 
% The optimization criteria usually revolve around safety-related factors such as contact impulse~\cite{ha2015multiple,kumar2017learning} or the descent of the center of mass~\cite{yun2014tripod}. 
% To simplify sequence tracking, some approaches utilize external objects like sticks or walls near the robot~\cite{tam2016fall,8460500}. 

% However, these methods share common drawbacks: they use simplified models to reduce computational costs, constrain the manageable fall direction to the sagittal plane, and contact sequence has limited expressive capacity to describe falls in dynamic tasks. 
% \guanya{we should emphasize these policies are non-autnomous, trigger by external signals, in stationary environments.}

% \guanya{Transitions from biped to quad: similarly, XXX} 
% Similarly, in the realm of quadrupedal robot, some works presented recovery policy after a fall \cite{ma2023learning, lee2019robust}. However, this does not reduce the damage caused during the falling process. Besides, some recent works have attached an arm to the robot body, which could help the robot avoid falling too severely and restore balance~\cite{ma2023learning, tang2023towards}. However, these methods rely on modifying the robot's hardware structure, potentially altering the task scope and reducing locomotion agility. 

\begin{figure}[t]
\centering     
        % \begin{subfigure}[h]{0.19\textwidth}
        %     \centering
        %     \includegraphics[height=2.5cm]{figs/reversed_mode.png}
        %     \caption{Reversed Mode}
        %     \label{reversed_mode}
        % \end{subfigure}\hspace{1mm}%
        % \begin{subfigure}[h]{0.09\textwidth}
        %     \centering
        %     \includegraphics[height=2.5cm]{figs/stand_pose.png}
        %     \caption{Standing}
        %     \label{stand_mode}
        % \end{subfigure}\hspace{1mm}%
        % \begin{subfigure}[h]{0.19\textwidth}
        %     \centering
        %     \includegraphics[height=2.5cm]{figs/regular_mode.png}
        %     \caption{Regular Mode}
        %     \label{regular_mode}
        % \end{subfigure}
        \includegraphics[width=0.8\linewidth]{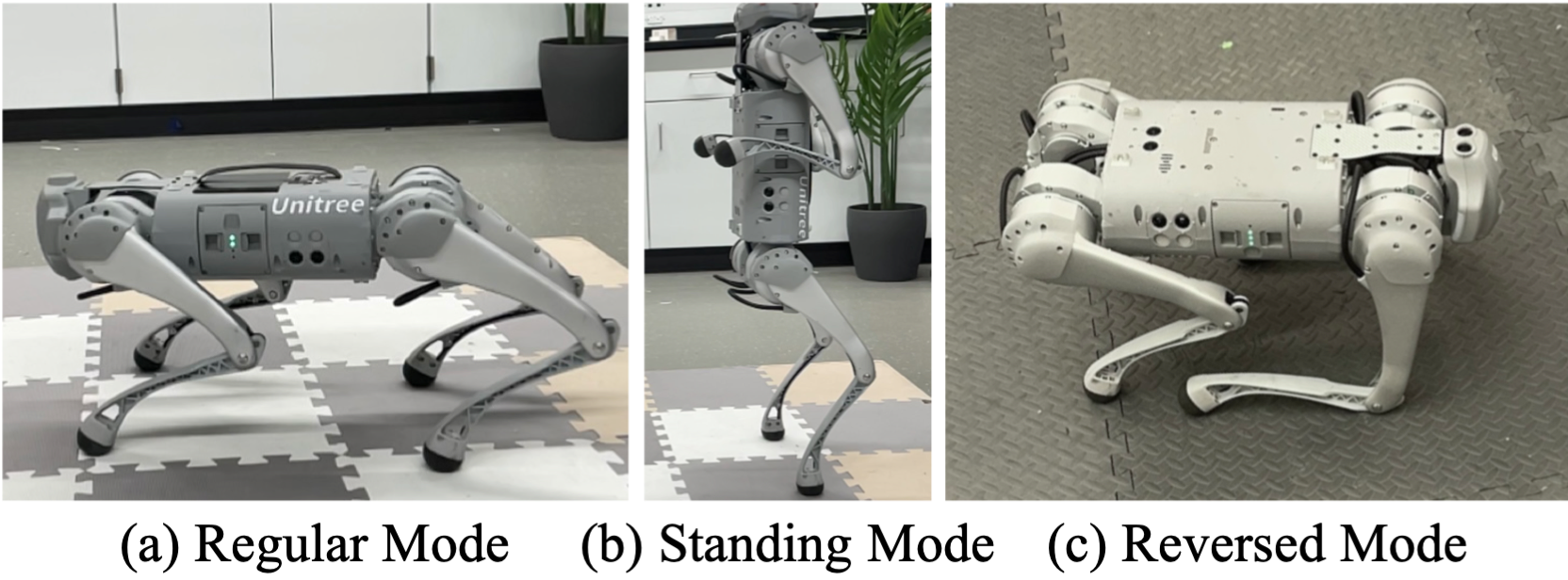}
\caption{The three stable modes of a quadrupedal robot.}
\vspace{-0.2in}
\label{stable_mode}
\end{figure}

% \guanya{make use of three stable modes for what? need to introduce our goal first} 
In this work, we seek to tackle the challenge of guaranteeing safe falls for quadrupeds during agile movements. We propose the \textbf{G}uardians as \textbf{Y}ou \textbf{F}all (\method) learning and control framework, which can autonomously detect unstable modes and actively initiate the safe falling procedure.  
We define three stable modes of the quadrupedal robot, as illustrated in Fig~\ref{stable_mode}, namely, the standing mode, regular mode, and reversed mode. \method leverages transitions between stable modes to address safety during dynamic tasks, drawing inspiration from animals' fall behaviors and the concept of attraction regions in nonlinear dynamical systems~\cite{kim2014optimal}. When animals fall with high initial momentum, they often roll on the ground to mitigate potential injuries without exerting much force. Similarly, the fundamental idea of our approach is that when perturbed, the robot may actively exit the attraction region of the initial stable mode and converge to another target stable mode in a safe and controllable manner, rather than passively leaving the initial mode and shifting to irrecoverable poses.

\begin{figure*}[t]
\begin{subfigure}{\textwidth}
%\hsize=\textwidth
        \centering
        \begin{minipage}{0.16\textwidth}
            \includegraphics[width=1\linewidth]{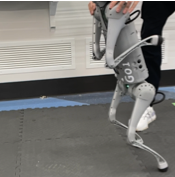}
        \end{minipage}
        \hspace{-6pt}
        \begin{minipage}{0.16\textwidth}
            \includegraphics[width=1\linewidth]{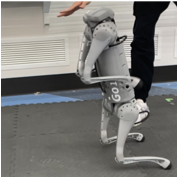}
        \end{minipage}
        \hspace{-6pt}
        \begin{minipage}{0.16\textwidth}
            \centering
            \includegraphics[width=1\linewidth]{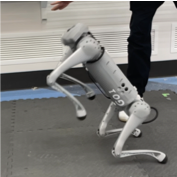}
        \end{minipage}
        \hspace{-6pt}
        \begin{minipage}{0.16\textwidth}
            \centering
            \includegraphics[width=1\linewidth]{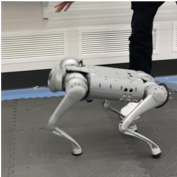}
        \end{minipage}
        \hspace{-6pt}
        \begin{minipage}{0.16\textwidth}
            \includegraphics[width=1\linewidth]{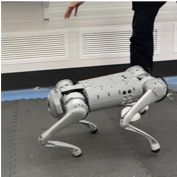}
        \end{minipage}
        \hspace{-6pt}
        \begin{minipage}{0.16\textwidth}
            \includegraphics[width=1\linewidth]{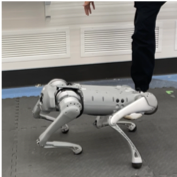}
        \end{minipage} 
    \caption{Safe falling to the reversed mode. In the reversed mode, the robot starts in a backward-leaning position.}
    \label{back_snapshots}
    \end{subfigure}
    \begin{subfigure}{\textwidth}
    \centering
        \begin{minipage}{0.16\textwidth}
            \includegraphics[width=1\linewidth]{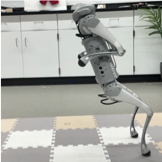}
        \end{minipage}
        \hspace{-6pt}
        \begin{minipage}{0.16\textwidth}
            \includegraphics[width=1\linewidth]{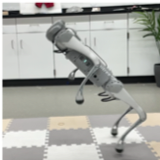}
        \end{minipage}
        \hspace{-6pt}
        \begin{minipage}{0.16\textwidth}
            \centering
            \includegraphics[width=1\linewidth]{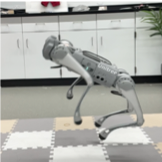}
        \end{minipage}
        \hspace{-6pt}
        \begin{minipage}{0.16\textwidth}
            \centering
            \includegraphics[width=1\linewidth]{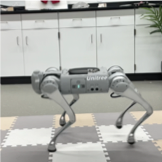}
        \end{minipage}
        \hspace{-6pt}
        \begin{minipage}{0.16\textwidth}
            \includegraphics[width=1\linewidth]{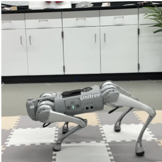}
        \end{minipage}
        \hspace{-6pt}
        \begin{minipage}{0.16\textwidth}
            \includegraphics[width=1\linewidth]{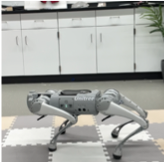}
        \end{minipage}
    \caption{Safe falling to the regular mode. In the regular mode, the robot starts in a forward-leaning position.}
    \label{forward_snapshots}
    \end{subfigure}
    \caption{Visualization of the safe falling procedure to the \textit{Reversed mode} and the \textit{Regular mode}.}
    \vspace{-0.15in}
\end{figure*}

% \guanya{Our goal is: guardian working controller with switching controller and recovery controller, such that the robot can safely recover to the regular mode.}

Our proposed \method utilizes a hierarchical architecture with a high-level planner scheduling three types of low-level policies, including a \emph{transition controller} for safe falling, a \emph{recovery controller} for recovery to the regular initialization posture, and a \emph{working policy} for performing highly dynamic tasks. 
\method can guard the robot's safety during the execution of the working policy, ensuring safety even if it falls and allowing it to return to the regular mode through the use of the transition controller and recovery controller.

% We refer to the policy that controls the robot to perform dynamic tasks as the \emph{working policy}. 

% We aim to train a controller that can adaptively transition between stable modes when the robotic dog is in an unstable state \guanya{mode}, facilitating the regaining of balance. This controller, named the \emph{transition controller}, plays a critical role in achieving this objective. If the converged state is in the reversed mode, we introduce a \emph{recovery controller} to transition the robot from the reversed mode to the regular mode. To enable the robot to effectively utilize these controllers, it must be capable of detecting unstable states and activating the appropriate controller when necessary. To automate this process, we trained a \emph{planner} to intelligently select between the transition controller, recovery controller, and the predefined working policy. %\guanya{no predefine}

% These components are seamlessly integrated into a unified framework that governs the robot's behavior according to the working policy while enabling smooth transitions to the transition controller whenever the robot enters an unstable state. In this work, our chosen working policy directs the quadrupedal robot to stand on two hind legs. \guanya{why don't we just define the working policy as a standing policy in early paragraphs?}

In summary, our contributions are as follows:
\begin{itemize}
\item We present \method, a hierarchical framework that ensures autonomous and active safe falling and recovery of quadrupeds when performing highly dynamic tasks.
\item We introduce a novel transition controller to mitigate falling damage through adaptive stable mode transitions.

% \item We propose three stable modes for quadrupeds: the regular mode, the reversed mode, and the standing mode as illustrated in Fig.~\ref{stable_mode}. \guanya{don't think it is a contribution. The first contribution should be proposing a new framework GAF, xxx}

\item In simulation and real-world experiments, \method reduces the maximum acceleration and jerk by $20\%$ $\sim$ $73\%$ compared to baselines.%\guanya{In simulation and real-world experiments, ...}

\end{itemize}

%% file: related.tex
\section{RELATED WORK}
\label{sec:related_work}

\subsection{Dynamic Control of Quadrupedal Robots}
Advancements in learning and control have equipped quadrupedal robots to perform dynamic tasks \cite{kumar2021rma, tan2018sim, margolis2023walk}, such as standing \cite{smith2023learning, yu2023language}, jumping \cite{park2015online, xie2020allsteps, margolis2021learning, yang2023cajun}, running \cite{DBLP:journals/corr/abs-1909-06586}, dribbling \cite{ji2023dribblebot} and parkour \cite{zhuang2023robot, caluwaerts2023barkour, cheng2023extreme}. 
Utilizing a linearized model, optimization techniques allow these robots to swiftly traverse rugged terrains \cite{DBLP:journals/corr/abs-1909-06586}. 
However, the reliability is related to model accuracy, and the robustness is guaranteed in states close to stable states. 
% Reinforcement learning (RL) offers a solution for dynamic tasks with nonlinear models. 
% For example, \cite{A1GoalkeepingHuang2022} uses hierarchical RL to train quadrupedal robots as goalkeepers to save balls by agile jumping. 
% Learning from real animal data also helps robots gain agile motions \cite{RoboImitationPeng20,vollenweider2023advanced, escontrela2022adversarial}.
% While a quadruped's stable state is typically all four feet on the ground, recent work has enabled quadrupeds to stand up from the ground and balance with only two hind feet, which is much more dynamic than prior works.
% \cite{smith2023learning} learns walking with two feet from experience transfer. 
% \cite{yu2023language} uses LLM to generate reward parameters for walking tasks and get low-level action by optimizing the reward using MPC. 
% \cite{cheng2023legs} uses two feet to stand and one foot to lean on the wall for balancing. 
% However, increased dynamism results in a higher risk of damage, and current research hasn't fully addressed safety measures for potential falls.
RL provides a solution for handling dynamic tasks with nonlinear models. For instance, in \cite{A1GoalkeepingHuang2022}, hierarchical RL is utilized to train quadrupedal robots as goalkeepers, enabling them to save balls through agile jumping. Learning from real animal data has also proven beneficial in helping robots achieve agile motions \cite{RoboImitationPeng20,vollenweider2023advanced, escontrela2022adversarial,wang2023amp}. While a quadruped's stable state traditionally involves having all four feet on the ground, recent advancements have allowed quadrupeds to stand up and maintain balance with only two hind feet, showcasing a higher level of dynamism compared to earlier works. In \cite{smith2023learning}, walking with two feet is learned through experience transfer. In \cite{yu2023language}, LLM is leveraged to generate reward parameters for walking tasks, and low-level actions are obtained by optimizing the reward using MPC. Additionally, \cite{cheng2023legs} demonstrates standing on two feet and leaning on a wall with one foot for balancing. However, this increased dynamism comes with a heightened risk of damage, and current research has not adequately addressed safety measures to mitigate potential falls.

\begin{figure*}[t]
\centering
\includegraphics[width=0.8\textwidth]{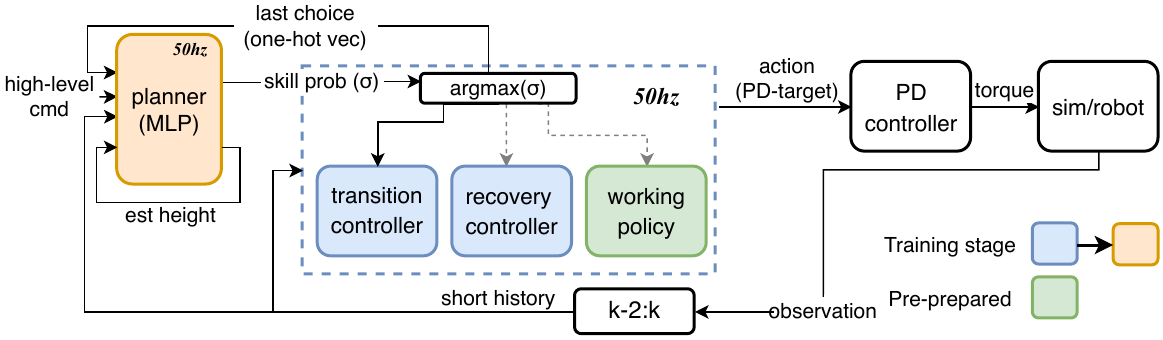}
\caption{Overview of our proposed \textbf{G}uardians as \textbf{Y}ou \textbf{F}all (GYF) framework. \method utilizes a hierarchical policy architecture to enable the quadrupedal robot's safe falling and recovery. \method consists of a high-level planner and three low-level policies, including the transition controller, recovery controller and the working policy.
% \guanya{need a high-level description in the caption} 
%\guanya{why only one arrow from argmax?}
}
\vspace{-0.2in}
\label{overview}
\end{figure*}

\subsection{Legged Robot Falling and Recovery}
Previous studies have primarily focused on investigating safe falling mechanisms for bipedal robots, employing optimization methods. In \cite{8460500}, a simplified robot model and a precomputed database of hand contact optimization are utilized to achieve real-time falling control with hand contact. Another approach presented in \cite{ha2015multiple} involves planning a contact sequence in an unstable initial state to dissipate the initial momentum. In \cite{yun2014tripod}, a swing foot and two hands are used to prevent the robot from falling completely to the ground. Additionally, \cite{kumar2017learning} adopts a mixture of actor-critic architecture to select the next contact body part and output the control action, aiming to minimize the maximum impulse.
Some works, such as \cite{samy2015falls, samy2017qp}, utilize posture reshaping to avoid singularity and implement adaptive gain compliance to reduce joint damage. Another strategy, presented in \cite{goswami2014direction}, involves altering the fall direction during the descent using foot placement and inertia reshaping.
More recently, researchers have explored the use of RL to train a recovery controller, aiming to restore the robot from an abnormal posture to its working posture, as seen in \cite{pmlr-v205-margolis23a, DBLP:journals/corr/abs-1901-07517}. However, these studies have not thoroughly addressed the safety of the falling process, particularly in highly dynamic scenarios.

%% file: method.tex
\section{METHOD: GUARDIANS AS YOU FALL}

In this section, we first state the safe falling problem of quadrupedal robots in Sec.~\ref{sec:problem_formulation} and present our proposed method \textbf{G}audians as \textbf{Y}ou \textbf{F}all (\method) in Sec.~\ref{sec:method_overview}. We then introduce three main modules of our method: transition controller in Sec.~\ref{sec:transition}, recovery controller in Sec.~\ref{sec:recovery}, and high-level planner in Sec.~\ref{sec:planner}.

%%%%%%%%%%%%%%%%%%%%%%%%%%%%%%%%%%%%%%%%%%%%%%%%%%%%%%%%%%%%%
\subsection{Problem Formulation: Safe Falling of Quadrupedal Robot}
\label{sec:problem_formulation}
This paper aims to address the challenge of ensuring safe falls for quadrupedal robots during agile movements. 
When performing such dynamic motions, quadrupedal robots are at risk of falling from high-energy states, as depicted in the leftmost figures in Fig.~\ref{back_snapshots} and Fig.~\ref{forward_snapshots}. 
Our objective is to introduce a method that not only mitigates the contact forces, joint torques, and motor jerks during a fall but also ensures the quadrupedal robot lands in a position optimal for swiftly resuming agile activities.

\textbf{Notations.} A quadrupedal robot's state comprises of the global position $\mathbf{p} \in \mathbb{R}^3$, linear velocity $\mathbf{v} \in \mathbb{R}^3$, body orientation $\mathbf{\Theta} \in \mathbb{SO}(3)$, angular velocity $\mathbf{\omega} \in \mathbb{R}^3$, joint position $\mathbf{q} \in \mathbb{R}^{12}$ and joint velocity $\mathbf{\dot{q}} \in \mathbb{R}^{12}$. 
% We use $\mathbf{s}=[\mathbf{p} ; \mathbf{v} ; \mathbf{\Theta} ; \mathbf{\omega} ; \mathbf{q} ; \mathbf{\dot{q}}]$ to denote the state. 
The robot's body height is $h = \mathbf{p}[2]$, and projected gravity is $\mathbf{g}\in \mathbb{R}^3$, which is a unit vector
capturing the robot’s orientation.
% To facilitate discussion, we define the $\emph{position}$ as $\mathbf{x}=[z ; \mathbf{\Theta} ; \mathbf{q}] $ to represent the state while solely considering morphological factors and disregarding velocity terms and xy-dimension of global position $\mathbf{p}$.

%%%%%%%%%%%%%%%%%%%%%%%%%%%%%%%%%%%%%%%%%%%%%%%%%%%%%%%%%%%%%
\subsection{Method Overview}
\label{sec:method_overview}

We propose \textbf{G}uardians as \textbf{Y}ou \textbf{F}all (\method), a hierarchical policy that actively adjusts the quadrupedal robot's configuration throughout the falling process. \method demonstrates dynamic falling behaviors, allowing quadrupedal robots to transition to novel stable modes beyond the standard upright stance, and actively tumble to regain stability.

\textbf{Stable modes.} We introduce three stable modes for a quadruped, illustrated in Fig.~\ref{stable_mode}. 
In dynamic scenarios, like when the robot is trying to stand upright, there's a potential for it to lean backward and lose equilibrium.
In such situations, we argue that returning to the typical upright stance as in Fig.~\ref{stable_mode}(a) is challenging.
Instead, it's more practical for the quadrupedal robot to land with its back oriented toward the ground as in Fig.~\ref{stable_mode}(c). 
% \guanya{the previous example is to motivate why we define three modes (not just standing and regular)?}
% . For safety concerns, we define three stable modes $\mathbf{x_{s}}$ 
%\guanya{standing?}
Hence, we define three types of stable modes of quadrupedal robots as follows. In \textbf{regular mode}, the quadrupedal robot stands with all four feet on the ground, belly facing downwards.
In \textbf{reversed mode}, the robot has all four feet on the ground but with its back facing downwards.
In \textbf{standing mode}, the robot is upright, balancing solely on its hind feet.
The regular and reversed modes are more stable than the standing mode in the presence of external forces.
% \begin{itemize}
%     \item \textbf{Regular mode}: The quadruped stands with all four feet on the ground, belly facing downwards.
%     \item \textbf{Reversed mode}: The quadruped has all four feet on the ground, but with its back facing downwards.
%     \item \textbf{Standing mode}: The quadruped is upright, balancing solely on its hind feet.
% \end{itemize}

% where all four feet are on the ground ensuring relative stability and safety: (1)Regular mode: standing with four feet on the ground; (2)Reversed mode: standing with four feet on the ground but upside-down.

\textbf{\method framework overview.} 
Our method, \method, illustrated in Fig~\ref{overview}, is developed to ensure the safe falling of quadrupedal robots.
\method utilizes a hierarchical architecture with \textit{a high-level planner}, which selects from low-level policies based on detected unstable states or a human-provided command.
The low-level policy set comprises the \textit{transition controller}, the \textit{recovery controller}, and other specialized task policies, denoted as working policies. 
These low-level policies generate joint-level PD targets, guiding the PD controller to produce joint torques.
Below, we delve deeper into the high-level planner and the low-level policies.
\begin{enumerate}
    \item The \textit{high-level planner} acts as a scheduler to transition between low-level policies to ensure a safe fall. 
    It estimates the height of robot's center of mass (CoM) and calculates the probability of selecting each low-level policy, considering both the robot's current state and human-provided commands. The low-level policy with the highest probability is activated.
    \item The \emph{transition controller} enables an adaptive shift from unstable postures to a suitable stable mode. Additionally, it can dynamically transition between stable modes, such as sideways rolling, before achieving full stabilization, thereby minimizing potential damage.
    \item Once stabilized, the \emph{recovery controller} facilitates the robot's transition from the reversed mode to the regular mode, ensuring the working policy can be initiated seamlessly.
    \item The working policies encompass task-specific actions like walking, standing, jumping, or climbing. It's crucial to note that our safe-falling method, \method, operates independently of these working policies.
\end{enumerate}

\textbf{Training pipeline.} \method utilizes a two-phase training pipeline. We assume that the working policies are pre-trained and frozen in our framework. In the first phase, we train the transition and recovery controllers individually. 
In the second phase, we freeze the transition and recovery controllers and train the planner via supervised learning to detect unstable postures and select low-level policies according to unstable posture detection and human-provided command.
All the low-level policies run at a frequency of 50Hz. 
We train low-level policies in simulation using NVIDIA Isaac gym simulator \cite{makoviychuk2021isaac} and legged gym codebase~\cite{rudin2022learning}. 
We use Proximal Policy Optimization (PPO) \cite{schulman2017proximal} to train the transition and recovery controllers. The actor and critic nets in PPO are Multi-Layer Perceptrons (MLPs) with hidden dims [512, 256, 128].  
The high-level planner also utilizes MLPs with hidden dims [512, 256, 128].
We let the PD controller's $K_p$ and $K_d$ be 20 and 0.5, respectively.

%%%%%%%%%%%%%%%%%%%%%%%%%%%%%%%%%%%%%%%%%%%%%%%%%%%%%%%%%%%%%
\vspace{-0.05in}
\subsection{Transition Controller}
\label{sec:transition}
\vspace{-0.05in}

The transition controller, denoted as $\pi_{tran}$, allows the quadrupedal robot to move from unstable positions either directly to a stable mode or dynamically between stable modes until it achieves a stable state. 
At each timestep $k$, it takes a short history containing three timesteps' information as input and outputs target joint angles, $\mathbf{a}_k = \pi_{tran}(\mathbf{o}_{k-2}, \mathbf{o}_{k-1}, \mathbf{o}_k)$.

\textbf{Observation and action spaces.}
The observation $\mathbf{o}_{k}$ is a 46-dimensional vector $\mathbf{o}_k = [\mathbf{\omega}_k, \mathbf{q}_k, \mathbf{\dot{q}}_k, \mathbf{g}_k, \mathbf{c}_k, \mathbf{a}_{k-1}]$ including angular velocity, joint positions, joint velocities, projected gravity $\mathbf{g}_k \in \mathbb{R}^{3}$, binary foot-contact states $\mathbf{c}_k \in \{0,1\}^4$, and last action $\mathbf{a}_{k-1} \in \mathbb{R}^{12}$.
The output action $\mathbf{a}_{k}$ consists of 12-dim target joint angles.

% To gain the ability to transition to and between stable positions and avoid the variance of training being too high, we design two kinds of settings:
% \begin{itemize}
%     \item Initialize with random configurations in the air. This helps to learn to transition from unstable positions to stable positions.
%     \item Initialize around regular mode or reversed mode, use flying balls weighs 10kg to hit the robot to render it unstable. The range of balls' velocity is $[-5, 5]m/s$. This helps to learn transitions between stable positions.
% \end{itemize}

% \emph{3)Reward functions:} 
\textbf{Rewards.} We use three types of rewards to train the controller in simulation, highlighted as follows. We present detailed reward expressions and weights on our website.
% The objective of the training is to enable the robotic dog to autonomously transition from unstable positions to stable positions or between different stable positions while ensuring safety. 
% The rewards can be categorized as follows:
\begin{itemize}
    % \mengdi{is body orientation part of the transition policy input? The axis of the orientation?}
    % \guanya{what is stable mode?} 
    \item \emph{Transition rewards:} 
    We define the nominal state of each stable mode containing joint angles $\mathbf{q}$, body height $h$ and projected gravity $\mathbf{g}$. 
    % Concretly, $[\mathbf{q}_{reg}, h_{reg}, \psi_{reg}]$  the regular mode and  $[\mathbf{q}_{rev}, h_{rev}, \psi_{rev}]$ for the reversed mode. 
    Based on the robot's current pitch angle, we encourage the robot to recover to a stable mode $m \in [\text{regular}, \text{reversed}]$. The reward is calculated based on the distance to the nominal states, as $r_{tran} = \alpha_q \| \mathbf{q} - \mathbf{q}_{m}\|_2 + \alpha_h\| h - h_m \|_2 + \alpha_{g}\| \mathbf{g} - \mathbf{g}_m \|_2 $, with weights $\alpha_q, \alpha_h, \text{ and } \alpha_{g}$. %\mengdi{Yikai: check pitch}
    % $\mathbf{s}_{reg}=[\mathbf{p} ; \mathbf{v} ; \mathbf{\Theta} ; \mathbf{\omega} ; \mathbf{q} ; \mathbf{\dot{q}}]$.
    % We encourage the robot to recover into one of the two available stable modes that is closest to its current posture. The target stable mode is selected based on the current body orientation, and rewards are given for bringing the joint positions, body height, and body orientation closer to the selected target stable mode.
    \item \emph{Safety rewards:} The safety reward $r_{safety}$ aims to reduce damage from collisions between the robot's rigid components (excluding its feet) and the flat ground. It penalizes the vertical component of contact force on rigid bodies, the vertical component of rigid bodies' net force, and the vertical component of the rigid bodies' yank, which is the change of net force. Rewarding small vertical components of the momentum change and forces helps encourage the robot to convert vertical momentum into horizontal momentum when touching the ground.
    \item \emph{Smooth rewards:} We penalize joint torques $\tau$ and the action changing rates to smooth the resulting motion. 
    % We care only about the vertical component because we assume the ground to be plane, and the vertical reaction forces of the ground cause the main damage. 
\end{itemize}

\textbf{Training setup.} At the beginning of each episode, we initialize the quadrupedal robot randomly in the air, around the regular mode, or around the reversed mode. Initializing in the air helps learn direct transitions from unstable postures to stable modes. Starting near the regular or reversed mode facilitates learning transitions between these stable modes.
To add external forces when the quadrupedal robot is positioned near the regular or reversed modes, we use a 10kg ball to strike it from the side. This ball has an initial velocity range of $[-5, 5]m/s$.

% \emph{4)Sim to real:}
\textbf{Sim-to-real transfer.}
We conduct experiments using the Unitree Go1 quadrupedal robot.
To account for hardware sensor inaccuracies and transmission delays, we incorporate observation noise at each timestep. 
To enhance the robustness and reduce the sim-to-real gap, we add randomization to certain physical factors, including the ground friction and restitution, $K_p$ and $K_d$ of the motors' PD controller, base mass, and the CoM position. 
Given that our safe-falling task typically spans just 0.2 seconds from unstable postures to stable modes, the prompt execution of actions becomes imperative.
To emulate hardware motor delays, we applied a consistent 20ms action delay in our simulations. 
Detailed specifications regarding noise and domain randomizations are available on our website.

%%%%%%%%%%%%%%%%%%%%%%%%%%%%%%%%%%%%%%%%%%%%%%%%%%%%%%%%%%%%%
\subsection{Recovery Controller}
\label{sec:recovery} 
After restoring the robot to its reversed mode, the recovery controller $\pi_rec$ is triggered upon receiving a specified user command. This transition guides the robot from the reversed mode to the regular mode, preparing it for the activation of the working policy.
$\pi_{rec}$ and $\pi_{tran}$ share identical observation and action spaces, as well as input and output formats.
The training procedure for $\pi_{rec}$ follows the procedure and the reward configurations of the recovery policy in \cite{pmlr-v205-margolis23a}. Specifically, the robot's position is initialized close to the reversed mode. Moreover, the recovery controller $\pi_{rec}$  is trained with the same domain randomization as the transition controller $\pi_{tran}$.

\subsection{High-level Planner}%\mengdi{Yikai: check}
\label{sec:planner}
% The \emph{high-level planner} selects from low-level policies based on current state or human command, so we generate target probability output based on predefined rules and train the planner to follow the predefined rules.

\textbf{Policy input and output.} At each time step $k$, the high-level planner $\pi_{planner}$ (1) selects a lower-level policy according to its probability output $q_k$ to enable safe falling, and (2) at the same time, outputs the estimated height $h^{est}_{k}$ of CoM to provide more comprehensive state information for accurate policy selection.
Concretely, $[\sigma_k, h^{est}_{k}] = \pi_{planner}(\mathbf{o}_{k-2}, \mathbf{o}_{k-1}, \mathbf{o}_k, \text{cmd}_k, h^{est}_{k-1}, \sigma_{k-1})$.
Here, $\sigma_k$ is a 3-dimensional one-hot vector representing the probability of each low-level policy. $h^{est}_{k}$ is the estimated CoM height at step $k$.
$\pi_{planner}$ takes four types of inputs, including the 3-step history $\mathbf{o}_{k-2}, \mathbf{o}_{k-1}, \mathbf{o}_k$, the high-level command $\text{cmd}_k \in \{0, 1, 2\}$ where the integer 0, 1, 2 represents the working policy, transition controller and recovery controller, respectively, the estimated CoM height at the previous timestep $h^{est}_{k-1}$ and the policy probability at the precious timestep $\sigma_{k-1}$.
We select the low-level policy index as $\arg \max(\sigma_k)$.
% \begin{itemize}
% \item a 3-step history of the low-level skill;
% \item the high-level command represented by integer 0, 1, 2 for working policy, transition controller and recovery controller respectively;
% \item the estimated CoM height;
% \item the skill probability represented by the 3-dim one-hot vector.
% \end{itemize}
 
% After the action, an argmax is performed on the probability distribution, and the corresponding low-level skill is executed.

% The planner has two main functions:
% \begin{itemize}
%     \item Selecting the appropriate lower-level skill based on high-level instructions and the current state;
%     \item Estimating the center of mass height to provide more comprehensive state information for accurate skill selection.
% \end{itemize}
% For simplicity, we use a single neural network to accomplish both of these functions. 

% \textbf{Policy Structure.}
% The output action of the planner includes a one-hot vector with 3 entries representing the probability of choosing each low-level skill, and a scalar indicating the estimated CoM height. The observation is a concatenation of four parts: 

%%%%%%%%%%%%%%%%%%%%%%%%%%%%%%%%%%%%%%%%%%%%%%%%%%%%%%%%%%%%%
% \subsubsection{Environment setting}
\textbf{Data collection.} 
% \guanya{Didn't understand this paragraph. Need polishing} 
We train $\pi_{planner}$ with supervised learning based on data collected in the simulation. 
We perform a sequence of low-level policies in each rollout and record the target probability output $\hat{\sigma}$ at each timestep as the label.
% \guanya{where is $\hat{q}$ from? -> explained in the following logics} 
% \mengdi{not sure about the following sentence: The key is to distinguish the standing-up and falling-down processes and remain in the stable modes until receiving a high-level command to stand up.} 
The robot is initialized near the regular mode with command $\text{cmd}=0$ and target $\hat{\sigma}=[1.0, 0.0, 0.0]$ to select a working policy to transit the robot to the standing mode. After the robot stands up and stabilizes in a standing position for a certain period, it will be hit by incoming balls at different speeds. 
For robots whose body acceleration surpasses a threshold after being disturbed by the ball, its planner's target becomes $\hat{\sigma} = [0.0, 1.0, 0.0]$ to select the transition controller for a safe falling. After that, the input command of robots whose estimated height is lower than a threshold becomes $\text{cmd}=1$. 
If the robot stabilizes in the reversed mode, the input command of robots is $\text{cmd}=2$, and the target is $\hat{\sigma}=[0.0, 0.0, 1.0]$.
If the robot stabilizes in the regular mode, the trajectory terminates. 
When collecting data, we guarantee that the working policy activates only when the robot is not falling or in the reversed mode.

% \textbf{Selecting Criteria.}
% The target of skill selection is defined as follows:
% \begin{itemize}
% \item Selection target is recovery controller whenever high-level command is 2;
% \item While high-level command is not 2, selection target is recovery controller when high-level command is 1 or low-level skill is set to transition controller;
% \item If the target is not 1 or 2, the target is 0. 
% \end{itemize}
% The idea is, the working policy is able to activate only when its not falling nor in the reversed mode.

%% file: result.tex
\begin{table*}[ht]
\centering
\begin{tabular}{@{}c|ccc|ccc|ccc@{}}
\toprule
% \diagbox{Method}{Scenario}
         & \multicolumn{3}{c|}{\textbf{Fall from a stage}} & \multicolumn{3}{c|}{\textbf{ Hit by a ball}} & \multicolumn{3}{c}{\textbf{Dropped from random orientations}} \\
 & \makecell{peak contact \\force[N]} & \makecell{peak base\\ jerk[$m/s^3$]} &\makecell{ peak base \\net force[N]} & \makecell{peak contact \\force[N]} & \makecell{peak base\\ jerk[$m/s^3$]} & \makecell{ peak base \\net force[N]} & \makecell{peak contact \\force[N]} & \makecell{peak base\\ jerk[$m/s^3$]} & \makecell{peak base \\net force[N]} \\ \midrule
\method     & \makecell{$\bf{927.53}$\\$\pm1120.41$}  & 
\makecell{$\bf{3326.09}$\\$\pm1909.23$} & 
\makecell{$\bf{328.84}$\\$\pm187.39$} & 
\makecell{$\bf{768.83}$\\$\pm971.40$}   &
\makecell{$\bf{3045.31}$\\$\pm1859.27$}  & 
\makecell{$\bf{287.79}$\\$\pm190.91$}  & 
\makecell{$\bf{32.96}$\\$\pm72.82$}       & \makecell{$\bf{1031.60}$\\$\pm289.94$}     & \makecell{$\bf{100.36}$\\$\pm26.64$}    \\ \hline
Standing & \makecell{$1826.52$\\ $\pm1243.09$} & 
\makecell{$4661.36$\\ $\pm1976.24$} & \makecell{$435.66$\\ 
$\pm196.06$} & \makecell{$1898.42$\\ $\pm1216.88$}  & 
\makecell{$4710.29$\\ $\pm1805.00$}  & \makecell{$439.44$\\ $\pm168.42$}  &  \makecell{$1614.06$\\$\pm887.46$}     &  \makecell{$3859.92$\\$\pm1130.01$}     &  \makecell{$340.11$\\$\pm104.52$}    \\ \hline
Damping  & \makecell{$1485.66$\\ $\pm1210.20$} & \makecell{$4170.42$\\ $\pm1454.35$} & \makecell{$399.55$\\ $\pm143.07$} & \makecell{$1736.59$\\$\pm1390.58$}  & \makecell{$3831.76$\\$\pm1884.66$}  & \makecell{$359.36$\\$\pm191.56$}  & \makecell{$1172.53$\\$\pm1003.96$}     & \makecell{$3717.58$\\$\pm1118.49$}     & \makecell{$358.98$\\$\pm102.05$}    \\ \bottomrule
\end{tabular}
\caption{Simulation experiment results. Smaller values show better falling behaviors.}
\vspace{-0.15in}
\label{diff_sce}
\end{table*}

\section{EXPERIMENTAL RESULTS}

In this section, we first show the unique behaviors of our proposed \method in Sec.~\ref{sec:transition_exp} through analyzing phase plane plots. We then present the simulation experiments in Sec.~\ref{exp_sec}-Sec.~\ref{sec:ablation} and the real-world experiments in Sec.~\ref{sec:real_exp}.
% \guanya{phase plan plots? ans: should be phase plane}

\subsection{Stable Mode Transitions}
\label{sec:transition_exp}

\begin{figure}[t]
\centering     
\begin{subfigure}[h]{0.8\linewidth}
    \centering
    \includegraphics[width=\linewidth]{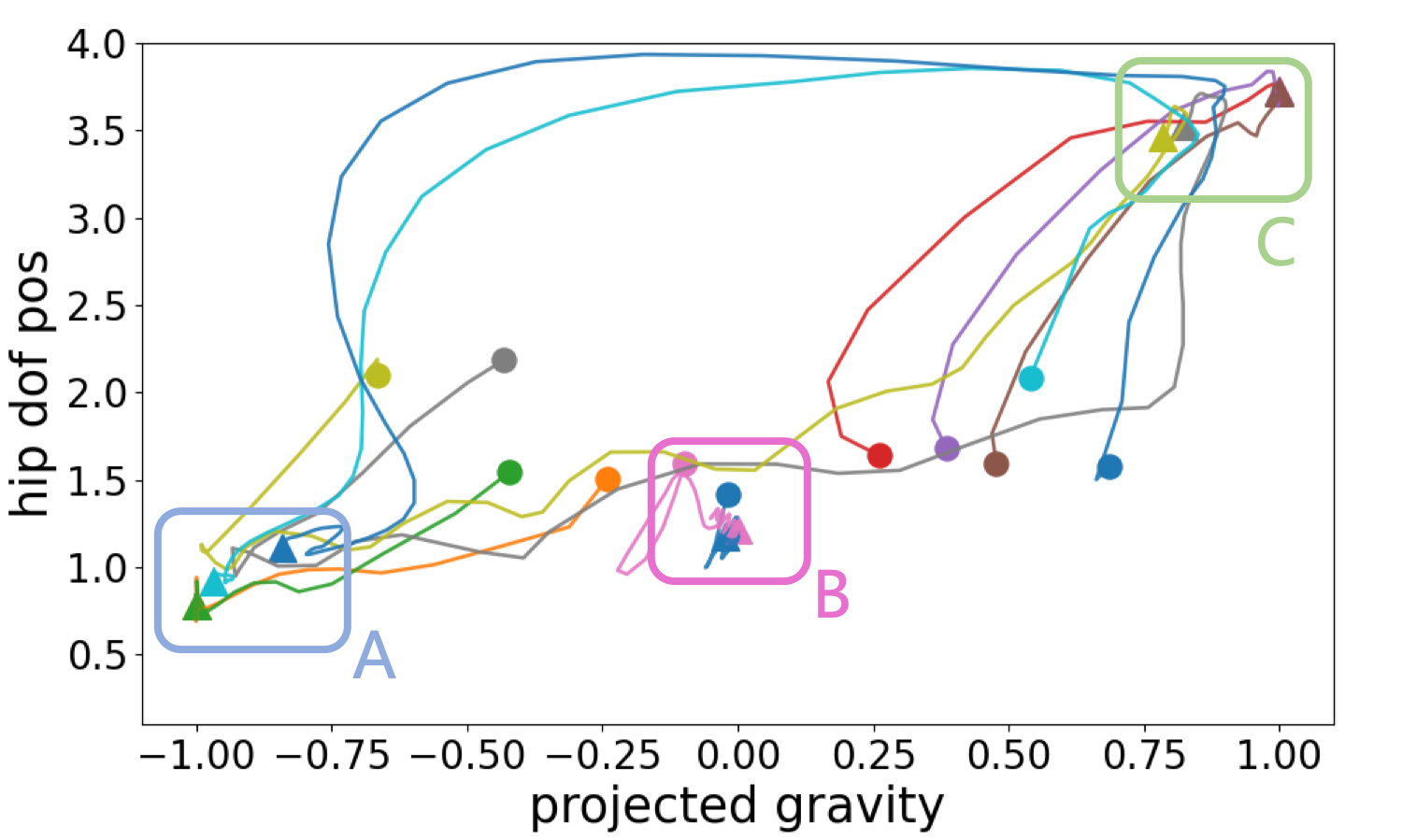}
    % \caption{\method (ours)}
    \label{transitions_init}
\end{subfigure}%
\\
\vspace{-0.3in}
\begin{subfigure}[h]{0.8\linewidth}
    \centering
    \includegraphics[width=\linewidth]{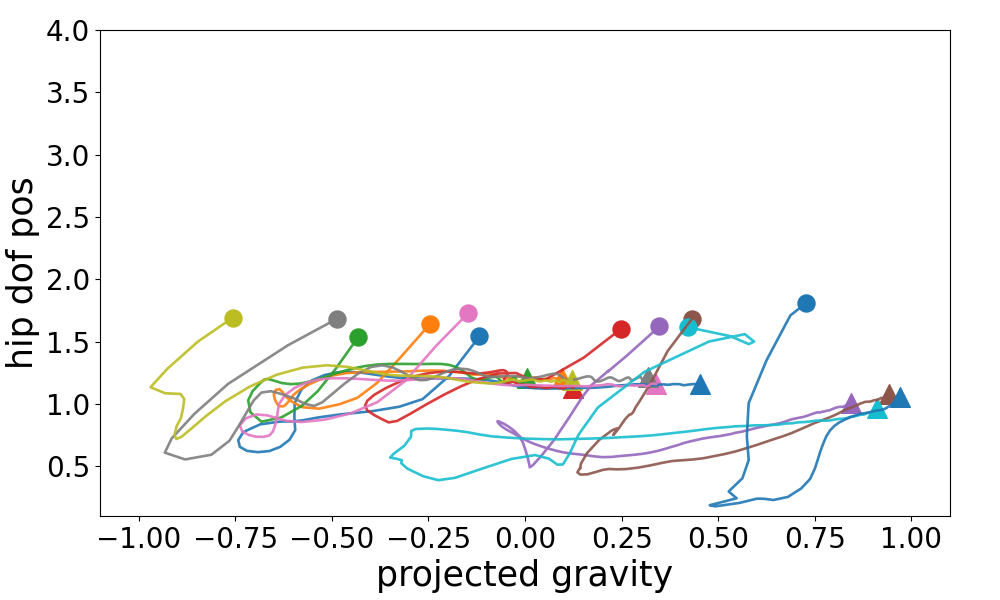}
    % \caption{Standing policy}
    \label{transitions_vel}
\end{subfigure}
\vspace{-0.2in}
\caption{Trajectories initialized from different states for both our proposed \method (\textbf{top}) and the standing policy (\textbf{bottom}). Each curve represents one trajectory, with red dots representing the starting point and the triangles representing the ending point. The attraction regions A, B, and C correspond to the regular, standing, and reversed modes, respectively. \method switches between different attraction regions, while the standing policy fails to discover the reversed mode.}
\vspace{-0.2in}
\label{transitions}
\end{figure}

We evaluate the capacity of our proposed \method for stable mode transitions in comparison with the standing policy, using phase plane plots. This provides insights into the distinct behaviors exhibited by \method during safe falling.

\textbf{Phase plane illustration.} We use two state variables to represent the state in the phase plane, including the projected gravity vector along the z-axis of the world frame, denoted as $\mathbf{g}_z$,
and the average of the robot's hip-thigh joint angles, denoted as $\Bar{\mathbf{q}}_{hip}$. 
$\mathbf{g}_z$ reflects the robot's deviation from the regular mode. It's more pertinent than projections on the x-axis or y-axis since the robot's rotation around the world frame's z-axis is less influential during safe falling.
In the regular mode $\mathbf{g}_z=-1$, in the reversed mode $\mathbf{g}_z=1$, and in the standing mode $\mathbf{g}_z\approx0$.
We choose $\Bar{\mathbf{q}}_{hip}$ since the maximum change in joint angles occurs in hip-thigh joints at the regular and reversed mode. This makes $\Bar{\mathbf{q}}_{hip}$ an effective indicator of the robot's posture.

% $\mathbf{g}_z$ can be obtained by rotating the unit gravity vector [0,0,-1] using the robot base quaternion then getting its z-component. 
% \begin{itemize}
%     \item Projection of the rotated gravity vector on z-axis of world frame, abbreviated as "proj-grav-z.". It can be obtained by rotating the unit gravity vector [0,0,-1] using the robot base quaternion then getting its z-component. This variable can be used to reflect the robot's deviation from vertical position and the robot's rotation along the z-axis is not important. It is more intuitive than quaternions and does not have the singularity issues of RPY angles. For example, at regular pose, the proj-grav-z is -1, at reversed pose it's 1, while at standing position it's close to 0.
    
%     \item Average of the robot's hip-thigh joint angles. At regular and reversed pose, the maximum change in joint angles occurs in these joints. So it can be used to reflect the robot's pose clearly.
% \end{itemize}

Based on simulation results, we visualize the phase plot curves in Fig.~\ref{transitions}. 
We draw multiple trajectories initialized from different states.
Each trajectory averages 5000 roll-outs from the same initial state, with a duration of 3 seconds. 
% Every red dot represents the trajectory's initialization state. 
% Each triangle stands for a terminal state. 
% We also highlight the attraction regions for the three stable modes, with attraction regions A, B, and C corresponding to the regular, standing, and reversed modes, respectively.
% Trajectories converge into three distinct attraction regions, denoted by A, B, and C. 

\textbf{Trajectory analysis.} There are three types of trajectories of our proposed \method as shown in Fig.~\ref{transitions}[top], including% \guanya{please fix the reference issue}
\begin{itemize}
    \item Initializing inside the standing mode's attraction region, and stabilizing in the standing mode (Region B);
    \item Initializing outside attraction regions and directly stabilizing in regular or reversed mode (Region B or C);
    \item Initializing outside attraction regions, first entering the regular mode and finally stabilizing in the reversed mode (Region A$\rightarrow$C), such as the yellow curve, or first entering the reversed mode and finally stabilizing in the regular mode (Region C$\rightarrow$A), such as the cyan curve. Such trajectories result from the high momentum when approaching the first stable mode, making the robot transit to the other one. 
\end{itemize}

% initializing outside the attraction, and be drawn to the attraction region of a stable mode. But the momentum is still too high to stabilize, so it continues to transition to another stable mode. 

% In Fig~\ref{transitions_init}, the roll-outs are obtained using the whole framework and have no initializing velocity, thus the paths whose initializing states are near the standing position end up near the standing position. Closer means a higher rate of roll-outs are ended up standing.

% In Fig~\ref{transitions_vel}, the roll-outs are initialized from different orientations with velocities ranging from -5m/s to 5 m/s. Thus many paths show transitions between the two stable positions to slow down to return to a final stable states more smoothly.

% With the ability to proactively utilize a stable mode as the recovery stability target and adaptively transit between stable modes, the robot could be safe even if entering an unstable state. 
Hence, with our proposed \method, the robot could safely fall even if entering an unstable state. 
In contrast, once outside the attraction region, the standing policy often results in erratic movements and ends up in a disordered position with entangled legs due to gravitational pull (Fig.~\ref{transitions}[bottom]). %\guanya{which figure?}

\begin{table*}[t]
\centering
\begin{tabular}{@{}c|cc|cc|cc|cc@{}}
\toprule
 & \multicolumn{2}{c|}{$\mathbf{45^\circ}$} & \multicolumn{2}{c|}{$\mathbf{75^\circ}$} & \multicolumn{2}{c|}{$\mathbf{105^\circ}$} & \multicolumn{2}{c}{$\mathbf{135^\circ}$ }\\
 & \makecell{peak \\acc[$m/s^2$]}    & \makecell{peak \\jerk[$m/s^3$]}    & \makecell{peak \\acc[$m/s^2$]}    & \makecell{peak \\jerk[$m/s^3$]}    & \makecell{peak \\acc[$m/s^2$]}     &\makecell{peak \\jerk[$m/s^3$]}    &\makecell{peak \\acc[$m/s^2$]}    & \makecell{peak \\jerk[$m/s^3$]}    \\ \midrule
\method     & \textbf{20.2} & \textbf{1203.08} & \textbf{25.71}  & \textbf{1424.37} & \textbf{26.85} & \textbf{1423.46} & \textbf{17.23} & \textbf{1177.45} \\
Standing & 29.58 & 1734.56 & 34.78  & 1934.76 & 46.22  & 2567.05 & 32.34 & 1569.56 \\
Damping  & 40.53 & 2323.48 & 42.30 & 2543.44 & 44.23  & 2329.88 & 29.27 & 1667.79 \\ \bottomrule
\end{tabular}
\caption{Real world experiment results. Smaller values show better falling behaviors. }
\label{real_exp}
\vspace{-0.2in}
\end{table*}
%\guanya{explain IMU has a sensing upper bound in practice...}

\subsection{Experimental Setup in Simulation}

\label{exp_sec}
\textbf{Evaluation scenarios.} After training, we directly evaluate all methods in three different scenarios:
\begin{itemize}
    \item \textit{Fall from a stage}: The robot falls from the edge of a stage with a height of 0.4m while standing on the stage with hind feet.
    \item \textit{Hit by a ball}: When the robot is standing on its hind feet, it is suddenly hit by a flying ball. The ball, weighing 10kg, has a variable speed between 1 and 5 m/s and comes from random directions.
    \item \textit{Dropped from random orientations}: The robot starts with random angles and no initial velocity, so its lowest part is 0.1m above the ground.
\end{itemize}

\textbf{Baselines.} We compare \method with two baselines.
\begin{itemize}
    \item \textit{Standing} policy. It represents the basic standing policy without the framework for safe fall control. We follow the algorithm in \cite{smith2023learning} to train the standing policy.
    \item \textit{Damping} policy. After the high-level planner detects falling, instead of selecting the transition controller as in \method, the robot consistently operates in damping mode with $k_{p}$=0 and $k_{d}$=0.5 for each motor. This follows the baseline setting from~\cite{ma2023learning}.%\mengdi{Yikai:citation}
\end{itemize}

\textbf{Metrics.} We use three metrics to evaluate safety and report their maximum value when falling in simulation in Table~\ref{diff_sce}. Each value is an average of 5000 roll-outs. 
\textit{Contact force} is the total contact force on the robot's rigid parts except feet.
\textit{Base jerk} is the magnitude of the jerk of the robot base.
\textit{Base net force} is the magnitude of the net force of robot base $|acc_{base}*mass_{base}|$, where $acc_{base}$ is the base acceleration and $mass_{base}$ is the base mass. 
% \begin{itemize}
%     \item 
%     \item 
%     \item 
% \end{itemize}

\begin{figure}[t]
\centering     
        \begin{minipage}{0.24\textwidth}
            \includegraphics[width=1\linewidth]{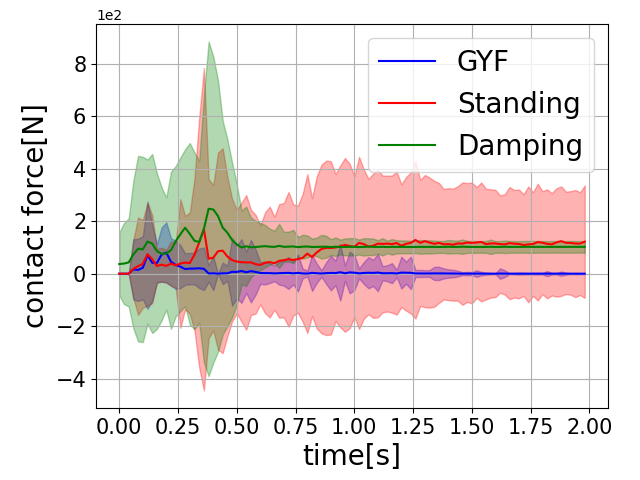}
        \end{minipage}
        \hspace{-6pt}
        \begin{minipage}{0.24\textwidth}
            \includegraphics[width=1\linewidth]{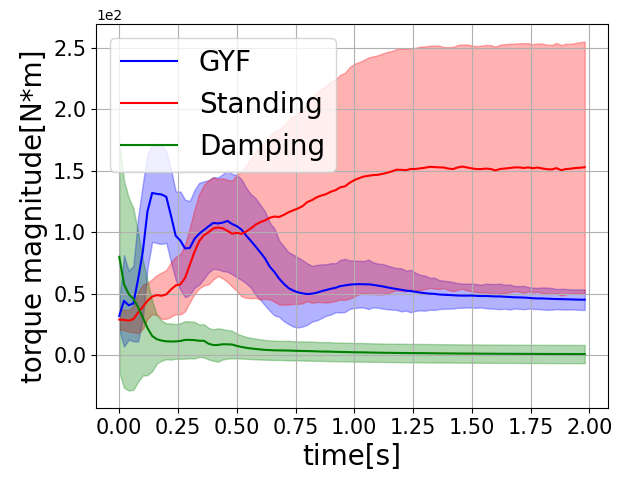}
        \end{minipage}
\caption{Comparison based on contact forces and torques in simulation. Each curve is averaged over 5000 roll-outs, and the shaded area represents the standard deviation.}
\vspace{-0.05in}
\label{fig:baseline}
\end{figure}

\begin{figure}[t]
\centering     
   
        \begin{minipage}{0.24\textwidth}
            \includegraphics[width=1\linewidth]{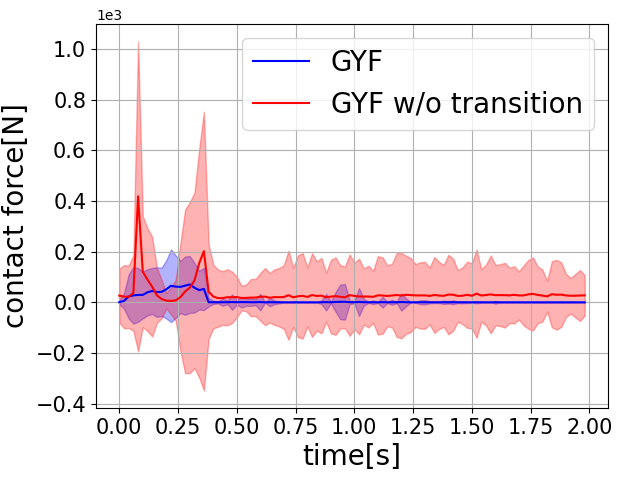}
        \end{minipage}
        \hspace{-6pt}
        \begin{minipage}{0.24\textwidth}
            \includegraphics[width=1\linewidth]{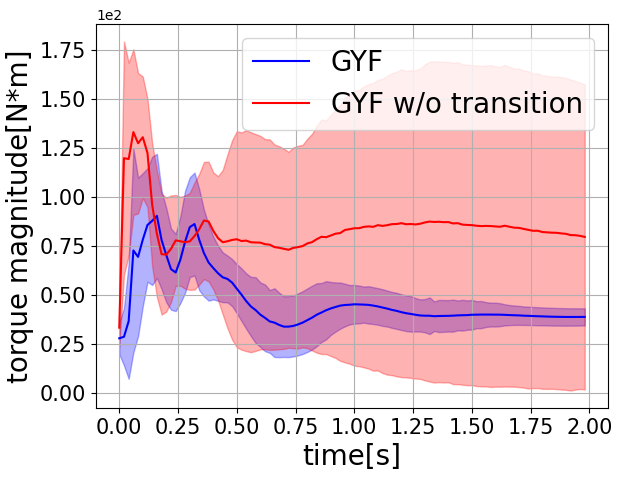}
        \end{minipage}
\caption{Ablation study about the transition between stable modes in simulation. Each curve is averaged over 5000 roll-outs, and the shaded area represents the standard deviation.}
\vspace{-0.2in}
\label{ablation}
\end{figure}

\subsection{Comparision with Baselines}
\label{sec:comparison}
We plot the contact force and motor torque of our proposed \method and baselines when being hit by a ball flying at a speed of 3m/s in Fig.~\ref{fig:baseline}.
The mean and variance of GYF's contact force are significantly lower than the baselines. GYF could select a stable mode for landing and gain a policy to transition between stable modes to reduce the contact forces. The standing policy cannot handle falls, making its motion chaotic in unstable periods, leading to a larger mean and variance for contact forces and torques. For the damping policy, the robot has no control over the motors, which makes its torque magnitude lower than \method, but still, it cannot handle falls, resulting in high contact forces. Due to the unpredicted motion of standing policy in unstable states, the mean and variance of its torque magnitude are also high at the initial stage.

\subsection{Ablation of Transitions Between Stable Modes}
\label{sec:ablation}
% \begin{figure}[htbp]
% \centering     
%         \begin{subfigure}[h]{0.25\textwidth}
%             \centering
            
%             \includegraphics[width=\linewidth]{wo_transition_contact.png}
%             \caption{}
%             \label{ablation-contact}
%         \end{subfigure}%
%         \begin{subfigure}[h]{0.25\textwidth}
%             \centering
%             \includegraphics[width=\linewidth]{wo_transition_net.png}
%             \caption{}
%             \label{ablation-net}
%         \end{subfigure}
%         \begin{subfigure}[h]{0.25\textwidth}
%             \centering
%             \includegraphics[width=\linewidth]{wo_transition_jerk.png}
%             \caption{}
%             \label{ablation-jerk}
%         \end{subfigure}%
%         \begin{subfigure}[h]{0.25\textwidth}
%             \centering
%             \includegraphics[width=\linewidth]{w_wo_transition_torque.png}
%             \caption{}
%             \label{ablation-torque}
%         \end{subfigure}
% \caption{The ablation of adaptively transition between stable positions. }
% \label{ablation}
% \end{figure}

We aim to demonstrate the role of transition between stable positions in reducing damage in highly dynamic conditions. Here, we train \textit{GYF w/o transition} in settings initializing with random configurations in the air but without being hit by a ball. Other training configurations are the same as the proposed policy. We test both policies in extremely dynamic conditions where the robot stands with two hind feet and is suddenly hit by a flying ball. Each ball weighs 10kg and flies at a speed of 3m/s from random directions. 
We present the main results in Fig.~\ref{ablation}. 
We record data after the ball hits the robot, then the ball will be removed. So, the plots do not include the direct influence of the ball hit.
The contact force curve of GYF w/o transition exhibits two distinct peaks, while the curve of GYF is relatively flat. That's because GYF could utilize transitions between the two stable modes to decrease momentum slowly and reduce impact. GYF w/o transition could leverage both stable positions but cannot transit between stable positions adaptively.
The plot of torque magnitude shows that the GYF does not rely on larger torques to regain balance forcefully.

\subsection{Real World Experiment}
\label{sec:real_exp}
We conducted real-world tests of our method to assess its benefits in safe falling and compare it to baselines. To quantify the performance, we first immobilize the joints of the robot dog in a 'standing' position and then posit it at different pitch angles relative to the regular mode. We then initialize the controller and let the robot fall. The maximum acceleration and maximum jerk during the falling process at different initial pitch angles are presented in Table~\ref{real_exp}. Each data of GYF is an average of 3 experiments. We only conduct one experiment with the baseline controller to prevent damage to the robot. In practice, we observed that the IMU integrated on the robot has a sensing upper bound. As a result, the actual experimental data for the baseline may exceed the values listed in the table. The results show that our proposed GYF has significantly smaller peak acceleration and peak jerks than baselines.

%% file: conclusion.tex
\section{Conclusion}
\label{sec:conclusion}

In this paper, we propose Guardians as You Fall (\method), which enables autonomous and active safe falling and recovery of quadrupeds in 0.2 seconds.
We define a novel stable mode as the reversed mode with quadruped's back facing the ground. 
\method is a hierarchical framework with a high-level planner scheduling low-level policies, including a transition controller that enables safe and agile falling behaviors, a recovery controller to facilitate fast restart, and working policies related to specific agile tasks. 
Our simulation and real-world experiments show that \method enables safe landing with contact forces and torques significantly smaller than baselines.
% We plan to integrate \method 
One interesting future work is to integrate additional explicit safety constraints in our framework.